\documentclass{article}

\usepackage{microtype}
\usepackage{graphicx}
\usepackage{subfigure}
\usepackage{booktabs} 

\usepackage{hyperref}
\usepackage[table]{xcolor}
\usepackage{wrapfig}

\usepackage[accepted]{icml2024}

\usepackage{amsmath}
\usepackage{amssymb}
\usepackage{mathtools}
\usepackage{amsthm}

\usepackage[capitalize,noabbrev]{cleveref}

\theoremstyle{plain}

\theoremstyle{definition}

\theoremstyle{remark}

\usepackage{booktabs}
\usepackage{multirow}

\usepackage[textsize=tiny]{todonotes}
\definecolor{Gray}{gray}{0.9}

\makeatletter
\newcommand{\algcolor}[2]{%
  \colorbox{#1}{\parbox{\dimexpr\linewidth-2\fboxsep}{#2}}%
}
\newcommand{\algemph}[1]{\algcolor{Gray}{#1}}
\makeatother

\newcommand{\NAME}[1]{LoCoCo}

\icmltitlerunning{LoCoCo: Dropping In Convolutions for Long Context Compression}

\begin{document}

\twocolumn[
\icmltitle{LoCoCo: Dropping In Convolutions for Long Context Compression}




\begin{icmlauthorlist}
\icmlauthor{Ruisi Cai}{ut}
\icmlauthor{Yuandong Tian}{meta}
\icmlauthor{Zhangyang Wang}{ut}
\icmlauthor{Beidi Chen}{meta,cmu}
\end{icmlauthorlist}

\icmlaffiliation{ut}{University of Texas at Austin}
\icmlaffiliation{meta}{Meta AI (FAIR)}
\icmlaffiliation{cmu}{Carnegie Mellon University}

\icmlcorrespondingauthor{Ruisi Cai}{ruisi.cai@utexas.edu}

\icmlkeywords{Machine Learning, ICML}

\vskip 0.3in
]



\printAffiliationsAndNotice{}  

\begin{abstract}
This paper tackles the memory hurdle of processing long context sequences in Large Language Models (LLMs), by presenting a novel approach, Dropping In Convolutions for \textbf{\underline{Lo}}ng \textbf{\underline{Co}}ntext \textbf{\underline{Co}}mpression (\textbf{\NAME{}}). \NAME{} employs only a fixed-size Key-Value (KV) cache, and can enhance efficiency in both inference and fine-tuning stages. Diverging from prior methods that selectively drop KV pairs based on heuristics, \NAME{} leverages a data-driven adaptive fusion technique, blending previous KV pairs with incoming tokens to minimize the loss of contextual information and ensure accurate attention modeling. This token integration is achieved through injecting one-dimensional \textbf{convolutional kernels} that dynamically calculate mixing weights for each KV cache slot.  Designed for broad compatibility with existing LLM frameworks, \NAME{} allows for straightforward ``drop-in" integration without needing architectural modifications, while incurring minimal tuning overhead. Experiments demonstrate that \NAME{} maintains consistently outstanding performance across various context lengths and can achieve a high context compression rate during both inference and fine-tuning phases. \underline{During inference}, we successfully compressed up to $3482$ tokens into a $128$-size KV cache, while retaining comparable performance to the full sequence - an accuracy improvement of up to $0.2791$ compared to baselines at the same cache size. \underline{During post-training tuning}, we also effectively extended the context length from 4K to 32K using a KV cache of fixed size 512, achieving performance similar to fine-tuning with entire sequences. Codes are available at: \url{https://github.com/VITA-Group/LoCoCo}.
\end{abstract}

\section{Introduction}

Large Language Models (LLMs) \citep{radford2018improving, radford2019language, brown2020language} excel across a variety of linguistic tasks, including text generation \citep{goyal2020evaluating, yuan2022wordcraft}, program synthesis \citep{chen2021evaluating, li2022competition}, question answering \citep{kamalloo2023evaluating}, and mathematical problem-solving \citep{lewkowycz2022solving}. These tasks typically involve processing extensive sequences, often requiring the analysis of thousands of tokens to derive outcomes based on comprehensive contextual information. For example, the task of summarizing extensive government reports, as seen in the GovReport section of SCROLLS \citep{shaham2022scrolls}, demands that LLMs efficiently sift through and distill key information from vast textual data, highlighting the need for models capable of handling long token sequences effectively.

Yet, transformers \citep{vaswani2017attention} struggle to process extensive token sequences due to their quadratic memory demands, which exceed the capacity of contemporary hardware. Attention computations are performed in blocks \citep{dai2019transformer}, with key and value states cached for subsequent encoding or decoding steps to mitigate this. However, this approach results in a Key-Value (KV) cache size that increases linearly with context length, quickly depleting GPU memory \citep{zhang2023h}. Recently,  StreamingLLM \citep{xiao2023efficient} attempted to reduce KV cache size by limiting each token's receptive field and incorporating "attention sinks". Concurrently, H\textsubscript{2}O \citep{zhang2023h} prunes tokens based on lower accumulated attention scores to stabilize KV cache size. Despite these efforts, both methods fail to leverage full-sequence information and adequately extend the context window. StreamingLLM's exclusion of all tokens in the context middle could significantly impair the model's ability to utilize the full long context (even completely ignore), known as ``lost in the middle"  \citep{liu2023lost}, while H\textsubscript{2}O struggles to extrapolate to longer sequences than the training context length \citep{han2023lm}.

Enhancing the context length in LLMs also necessitates increasing the block size during fine-tuning \citep{press2021train, chen2023extending}, introducing a significant memory challenge. While attention approximation methods like \cite{choromanski2020rethinking, kitaev2020reformer, xiong2021nystromformer} reduce training expenses, they do not alleviate memory demands at inference, as the KV cache continues to explode with longer predictions. Another representative work LongLoRA \citep{chen2023longlora} leveraged locally grouped attention alongside LoRA \citep{hu2021lora} for quick adaptation to longer-context data. However, implementing LongLoRA necessitates several modifications to the architecture of pre-trained LLMs for fine-tuning, hence not yet a hassle-free ``drop-in" option. Besides, the LongLoRA-tuned model may compromise its performance if we still use smaller context sizes (see our experiments).

In our study, we address the challenge of efficiently managing long contexts in \textbf{both inference and fine-tuning} phases. We introduce a novel method, 
\textbf{\underline{Lo}}ng \textbf{\underline{Co}}ntext Compressison by Dropping-In \textbf{\underline{Co}}nvolutions, abbreviated as \textbf{\NAME{}}. This technique employs a \textbf{static-size KV cache} for segment-level attention processes, ensuring peak memory usage remains unchanged. \NAME{} departs from traditional methods of dropping KV pairs based on pre-defined or ad-hoc rules \citep{zhang2023h, xiao2023efficient}, instead adopting a data-driven \textbf{adaptive fusion} approach that merges prior KV pairs with new tokens. This fusion minimizes the loss of the whole context and achieves accurate attention modeling. Specifically, \NAME{} utilizes one-dimensional \textbf{convolutional kernels} to calculate mixing weights for each KV cache slot, integrating incoming tokens efficiently \citep{kim2014convolutional, poli2023hyena, massaroli2023laughing}. This strategy is informed by the \textbf{insight} that autoregressive generation benefits from the continuity provided by shifting windows, and introducing the shift-invariant operation of convolutions can reinforce the sequence's stationary inductive bias. It counters potential discontinuities that might arise from excluding tokens during the generative process (e.g., one token in the middle might contribute to the current generation, but ``suddenly" be dropped when generating the next token). 

It is important to note that \NAME{} is designed to be \textbf{universally compatible} with existing LLM architectures, allowing for seamless integration without necessitating any modifications to the original model designs. It requires merely ``\textbf{dropping in}" a few extra convolution layers, incurs a small tuning overhead, yet can achieve consistently effective performance across various context lengths, with \textbf{high context compression rates} for both inference and fine-tuning.

Our contributions could be summarized as follows:\vspace{-0.7em}
\begin{itemize}
\item We introduced the novel (\NAME{}) method  to manage long contexts efficiently at both inference and fine-tuning, employing a static-size KV cache and data-driven adaptive fusion of context information.
\item \NAME{} utilized one-dimensional convolutional kernels for dynamic weight calculation in the KV cache, enhancing accurate attention modeling while addressing the challenges of sequence continuity and the stationary inductive bias for autoregressive generation.
\item \NAME{} highlights universal compatibility with existing LLM architectures, enabling easy ``drop-in" integration without extra design modifications, and achieving high context compression rates across different context lengths with minimal tuning overhead.
\item During inference, we successfully compress up to $3482$ tokens into a $128$-size KV cache, while retaining comparable performance to the full sequence - an accuracy improvement of up to $0.2791$ compared to baselines at the same cache size. During post-training tuning, we extended the context length from 4K to 32K using a KV cache of fixed size 512, achieving performance similar to fine-tuning with entire sequences.
\end{itemize}

\section{Related Work}

\subsection{Long-Context Inference}

Generating long contexts necessitates a KV cache for preceding tokens and incurs a significant memory overhead. For memory-efficient inference, \citet{zhang2023h} proposes mitigating KV cache demands during long-context generation through auto-regressive token eviction. Furthermore, \citet{ribar2023sparq} optimizes memory usage by selectively fetching from the cached history.
Approaching differently, \citet{jiang2023longllmlingua} focuses on prompt compression techniques to create concise yet expressive prompts. Meanwhile, \citet{xiao2023efficient} achieves infinite-length context generation by only storing tokens within a local window plus ``attention sink'' tokens, and rolling position embeddings.
However, they fall short of utilizing full-sequence information and extending the context window.

\subsection{Long-Context Fine-tuning}

The limited sequence length of pre-trained LLMs and their inability to handle long-context data effectively are major concerns for practitioners. To address this, strategies such as extending the context length through fine-tuning have been explored \citet{xiong2023effective}. The work of \citet{dai2019transformer} introduces a segment-level recurrence mechanism using fixed-length training segments. Other approaches include positional interpolation \cite{chen2023extending}, NTK-aware embedding \cite{ntk}, Yarn \cite{peng2023yarn}, positional skipping \cite{zhu2023pose}, self-extension \cite{jin2024llm}, stabilized attention entropy \cite{zhang2024extending}, and so on.
Additionally, landmark attention \cite{mohtashami2023landmark} introduces a gating mechanism based on landmark tokens, each representing a block of tokens. This method selectively retains ``landmarks'' in memory, utilizing other memory resources (e.g., CPU memory or disk) for storing the remaining tokens. \citet{tworkowski2023focused} employs contrastive learning, LongLoRA \cite{chen2023longlora} introduces shifted sparse attention and parameter-efficient fine-tuning. \citet{zhang2023efficient} investigates the necessity of attending to long-context tokens in a layer-wise manner.

\subsection{Attention Approximation}
Efforts to mitigate the quadratic complexity of transformers primarily focus on attention approximation. A comprehensive review of the rich literature can be found in \cite{tay2022efficient}. Specifically, \citet{child2019generating,kitaev2020reformer, roy2021efficient} leverages sparsity, and \citet{choromanski2020rethinking, katharopoulos2020transformers, wang2020linformer} utilizes low-rank approximation. \citet{beltagy2020longformer, zaheer2020big} approximated the full attention with both local and global attention.
Nevertheless, none of these approaches eliminate the memory bottleneck for the KV cache.

\subsection{Language Model Design with Built-In Convolutions}

\cite{dauphin2017language} introduced the first convolutional language model that rivaled strong recurrent models on large-scale language tasks. More recently, \cite{poli2023hyena, arora2023zoology} proposed using long convolutions to completely replace attention mechanisms in transformers. Additionally, state-space models (SSMs) can be computed as either convolutions or recurrences, achieving sub-quadratic training and constant inference complexity \cite{gu2021efficiently}. Architectures utilizing implicit convolutional filters \cite{poli2023hyena} can be converted to SSMs via a simple distillation step \cite{poli2023hyena, massaroli2023laughing}. These designs inspired our research; however, our work has a different focus of providing ``drop-in" components to enhance the long-context capability of pre-trained LLMs.

\begin{figure*}[tb]
    \centering
    \includegraphics[width=0.9\linewidth]{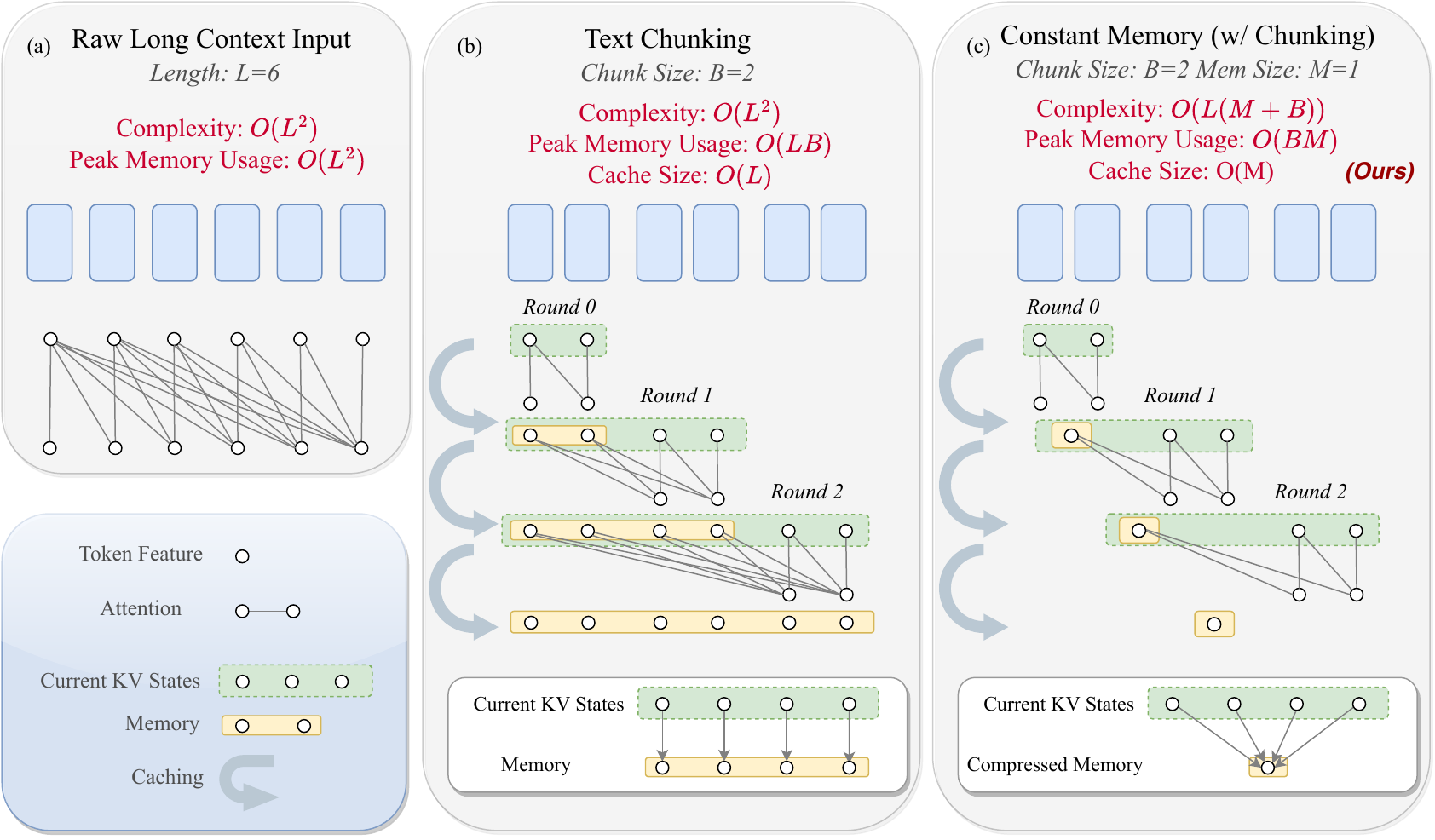}
    \caption{Overview of our pipeline. We process the long sequences block-wisely and maintain a fixed-size compressed memory.}
    \label{fig:Setting}
\end{figure*}

\section{Methodology}

\subsection{Segment-Level Attention with Long Sequences}

The attention mechanism \cite{vaswani2017attention} plays as a crucial component in transformers.
Suppose the sequence length is $L$ and the hidden dimension is $d$.
In causal language modeling, an attention block receives query, key, and value matrices $\boldsymbol{Q}, \boldsymbol{K}, \boldsymbol{V} \in \mathbb{R}^{d\times L}$, and computes outputs as:
\begin{align}\label{eqn:attention}
\operatorname{Attn}(\boldsymbol{K}, \boldsymbol{Q}, \boldsymbol{V}) = \boldsymbol{V} \mathrm{softmax}\left(\boldsymbol{K}^\top \boldsymbol{Q} \odot \boldsymbol{M} / \sqrt{d} \right),
\end{align}
where $\boldsymbol{M}$ is a lower-triangular causal mask.
Intuitively, causal attention only allows tokens to aggregate information from past tokens.
Given a sequence with $L$ tokens $\boldsymbol{X} = \begin{bmatrix} \boldsymbol{x}_1 & \cdots & \boldsymbol{x}_{L} \end{bmatrix} \in \mathbb{R}^{d \times L}$, the query, key, value matrices are computed as the linear projections of $\boldsymbol{X}$:

\begin{align}\label{eqn:qkv_proj}
\boldsymbol{K} = \boldsymbol{W}_K\boldsymbol{X}, \boldsymbol{Q} = \boldsymbol{W}_Q\boldsymbol{X}, \boldsymbol{V} = \boldsymbol{W}_V\boldsymbol{X}.
\end{align}
As illustrated in the Figure~\ref{fig:Setting} (a), acquiring full attention matrix $\mathrm{softmax}(\boldsymbol{K}^\top\boldsymbol{Q})$ requires $\mathcal{O}(L^2)$ peak memory cost.
When $L$ is large, i.e. handling long sequential data, full attention computation tend to run out of GPU memory rapidly.

Context chunking is a common practice for reducing peak memory usage during training.
Owing to the causality, Transformer-XL \citep{dai2019transformer} introduces a recurrent computation mechanism by caching and reusing the hidden states to extend the context length for both training and inference.
Specifically, the whole sequence is divided into a couple of segments, each then processed sequentially.
The intermediate key and value states will be stored in the memory.
Previously cached KV pairs will be used for computing the token representation in the subsequent segments.

Algorithm \ref{alg:seg_attn} presents the detailed procedure for the training-time attention computation.
Suppose the input sequence of length $L$ can be divided into $N$ segments, where each block has $B$ tokens, i.e. $L = NB$.
\begin{algorithm}[t]
\caption{Segment-level Attention (Training Time)}
\label{alg:seg_attn}
\begin{algorithmic}
\STATE {\bf Input}: A full sequence of length $L$: $\boldsymbol{x}_1, \cdots, \boldsymbol{x}_L$, block size $B$, the number of segments $N$.
\STATE Initialize an empty cached KV pairs as $\widetilde{\boldsymbol{K}}$, $\widetilde{\boldsymbol{V}}$.
\FOR{$n = 1, \cdots, N$}
\STATE {\bf Step 1} - Let $\boldsymbol{X}_n = \begin{bmatrix} \boldsymbol{x}_{nB} & \cdots & \boldsymbol{x}_{(n+1)B - 1} \end{bmatrix} \in \mathbb{R}^{d \times B}$ collect a sequence of the $n$-th segment.
\STATE {\bf Step 2} - Calculate key, query, values: $\boldsymbol{Q}_n = \boldsymbol{W}_Q \boldsymbol{X}_n$, $\boldsymbol{K}_n = \boldsymbol{W}_K \boldsymbol{X}_n$, and $\boldsymbol{V}_n = \boldsymbol{W}_V \boldsymbol{X}_n$.
\STATE {\bf Step 3} - {Perform attention as: \\ \hspace{3em} $\boldsymbol{O}_n \leftarrow \operatorname{Attn}([\widetilde{\boldsymbol{K}}, \boldsymbol{K}_n], \boldsymbol{Q}_n, [\widetilde{\boldsymbol{V}}, \boldsymbol{V}_n])$}
\STATE {\bf Step 4} - Update cached KV pairs: \\ \hspace{3em} $\widetilde{\boldsymbol{K}} \leftarrow [\widetilde{\boldsymbol{K}}, \boldsymbol{K}_n]$,  $\widetilde{\boldsymbol{V}} \leftarrow [\widetilde{\boldsymbol{V}}, \boldsymbol{V}_n]$.
\ENDFOR
\STATE {\bf Return} $\begin{bmatrix}
\boldsymbol{O}_1 & \cdots & \boldsymbol{O}_N
\end{bmatrix}$
\end{algorithmic}
\end{algorithm}
The symbol $[\cdot, \cdot]$ therein denotes the concatenation of two matrices' columns.

Note that auto-regressive generation is a special case of segment-level attention at $B=1$.
This is, tokens come in sequel and attention is only computed between the incoming query and past KV pairs.
The cached KV pairs $\widetilde{\boldsymbol{K}}, \widetilde{\boldsymbol{V}}$ are known as \textit{KV cache} for short in the inference mode \footnote{With a slight ambiguity, we also refer to the training-time saved KV pairs as the KV cache since their functionality is identical to their test-time counterparts.}.

As illustrated in Figure~\ref{fig:Setting}(b), by performing context chunking, the memory used by attention for the $r$-th block is $\mathcal{O}(B^2r)$, and the memory to store past KV pairs is $\mathcal{O}(Br)$.
Hence, the peak memory usage for computing the full attention is reduced to $\mathcal{O}(LB)$, which occurs at the ${N}$-th round when the last token needs to attend all previous key and value blocks $\{\boldsymbol{K}_n, \boldsymbol{V}_n, n \in \{1,...,N\}\}$.
The deduction of peak memory usage comes at the cost of increased caching memory, which grows linearly with the sequence length.

\subsection{Convolution as a Context Compression Operator}

So far, the peak memory footprint has been reduced from quadratic to linear concerning sequence length. However, this linear growth of the KV cache can still lead to excessive memory usage as the sequence length increases \cite{zhang2023h}. Early attempts using k-NN lookup \cite{wu2022memorizing} and gating mechanisms \cite{mohtashami2023random} enable sparse token selection to save memory but still require caching all previous tokens, resulting in a cache size of $\mathcal{O}(L)$. In this section, we introduce a framework that further optimizes this linear complexity to a constant size.

Compressing past token information using a fixed-size hidden space is well-documented in the literature. Notably, State Space Models (SSMs) utilize a fixed-dimension latent vector to represent all prior tokens, showing great promise for long-sequence modeling \cite{gu2021combining, gu2021efficiently, gu2020hippo, gu2022parameterization, gupta2022diagonal, fu2022hungry, gu2023mamba}. This hidden vector interacts with incoming tokens on behalf of all previous tokens. 

Inspired by this, we propose allocating at most $M$ slots to store past KV pairs, allowing subsequent sequence blocks to attend to these compressed KV states. We replace the simple concatenation in Step 4 with a KV compression operator $\mathcal{C}$:
\begin{align}
\label{equ:compression_operator}
\widetilde{\boldsymbol{K}} \leftarrow \mathcal{C}([\widetilde{\boldsymbol{K}}, \boldsymbol{K}_n]), \quad \widetilde{\boldsymbol{V}} \leftarrow \mathcal{C}([\widetilde{\boldsymbol{V}}, \boldsymbol{V}_n]),
\end{align}
where $\mathcal{C}$ maps a longer sequence to a sequence of length $M$.
We next elaborate on our instantiation of $\mathcal{C}$.

\subsubsection{Convolutional Token Compressor}

There are various ways to implement the sequence function $\mathcal{C}$ to meet the above definition. In this paper, we propose modeling the update rule of the KV cache as a weighted fusion between existing cache entries and newly input tokens. Formally, for all $\forall i \in [M]$:
\begin{align}
\widetilde{\boldsymbol{k}}_i \leftarrow \sum_{j=1}^{B} w_{i,j} \boldsymbol{k}_j + \sum_{j=1}^{M} \widetilde{w}_{i,j} \widetilde{\boldsymbol{k}}_j  \label{eqn:token_merge_k} \\
\widetilde{\boldsymbol{v}}_i \leftarrow \sum_{j=1}^{B} w_{i,j} \boldsymbol{v}_j + \sum_{j=1}^{M} \widetilde{w}_{i,j} \widetilde{\boldsymbol{v}}_j,  \label{eqn:token_merge_q}
\end{align}
where $w_{i,j}$ denotes the contribution of the $j$-th token in the input block to the $i$-th entry in the cache, and similarly $\widetilde{w}_{i,j}$ the contribution of the $j$-th token in the existing cache to the $i$-th entry in the updated cache.
Here weights for keys and values are shared to preserve token correspondence.

We further identify three key properties desired for $\{w_{i,j}\}$ and $\{\widetilde{w}_{i,j}\}$:
\textit{\textbf{1) Efficiency}}: computing these weights is an intermediate step of performing attention, and hence its overheads should be negligible - otherwise we beat our purpose.
\textit{\textbf{2) Learnability}}: Ad-hoc $\{w_{i,j}\}$ and $\{\widetilde{w}_{i,j}\}$, such as averaging (i.e., uniform weights) or heuristic-based token dropping (i.e., many zero weights) \citep{zhang2023h}, may not be flexible enough or introduce extra bias (e.g., locality \citep{chen2023longlora} or ``lost in the middle" \citep{liu2023lost}). 
\textit{\textbf{3) Stationarity}}: the compression policy must be globally informed and stable concerning token position, ensuring that compressed KV states update continuously as tokens are processed. This addresses the potential disruptions in KV states caused by dropping tokens during the generative process \cite{zhang2023h}.

It has not escaped our notice that convolutional kernels fulfill all the aforementioned requirements. Therefore, we propose using convolutional layers to generate ${w_{i,j}}$ and ${\widetilde{w}_{i,j}}$. Specifically, a 1D convolution will process all the pairs existing in the KV cache and the newly incoming segment, assigning each token an $M$-dimensional output that indicates its importance for each slot in the updated cache. 
Formally, we denote $\boldsymbol{g}: \mathbb{R}^{2d \times (M+B)} \rightarrow \mathbb{R}^{M \times (M+B)}$ as a Convolutional Neural Network (CNN) with $2d$ input channels and $M$ output channels.
Then weights $\{w_{i,j}\}$ and $\{\widetilde{w}_{i,j}\}$ are computed as:
\begin{align}
\boldsymbol{W} &\leftarrow \boldsymbol{g} \circledast \begin{bmatrix}
\boldsymbol{K}_n & \widetilde{\boldsymbol{K}}  \label{eqn:token_weight} \\
\boldsymbol{V}_{n} & \widetilde{\boldsymbol{V}}
\end{bmatrix} \in \mathbb{R}^{M \times (M+B)}, \\
w_{i, j} &\leftarrow \frac{\boldsymbol{W}_{i,j}}{\sum_{k=1}^{M+B} \boldsymbol{W}_{i,k}}, \quad \widetilde{w}_{i, j} \leftarrow \frac{\boldsymbol{W}_{i,j+B}}{\sum_{k=1}^{M+B} \boldsymbol{W}_{i,k}}, \label{eqn:token_weight_norm}
\end{align}
where $\circledast$ denotes multi-channel convolution operation along columns of two operands.
Here we normalize the prediction from the CNN kernel $\boldsymbol{g}$ as the final blending weights.
Convolution parameters are trained end-to-end with a small set of calibration data.

We name our approach as \textit{Long Context Compression by Dropping-In Convolutions}, or \textbf{\NAME{}} for short.
We summarize the outline of \NAME{} in Algorithm \ref{alg:lc3_attn}, where the major differences from Algorithm \ref{alg:seg_attn} are \colorbox{Gray}{highlighted} in Steps 4 and 5.
When the number of KV entries to be stored $\#(\widetilde{\boldsymbol{K}}, \widetilde{\boldsymbol{V}}) + B$ surpasses the number of slots $M$, we apply convolution-based compression between cached entries and newly added KV pairs.
Otherwise, we preserve all KV pairs in the memory.
In our implementation, we adopt a shallow CNN for each attention layer. The convolutional head consists of a single convolution layer with kernel size $21$.
In addition, we prepend a ReLU as the activation function.

\begin{algorithm}[t]
\caption{\NAME\, Attention (Training Time)}
\label{alg:lc3_attn}
\begin{algorithmic}
\STATE {\bf Input}: A full sequence of length $L$: $\boldsymbol{x}_1, \cdots, \boldsymbol{x}_L$, block size $B$, the number of segments $N$, the number of total cached KV entries $M$.
\STATE Initialize an empty cached KV pairs as $\widetilde{\boldsymbol{K}}$, $\widetilde{\boldsymbol{V}}$.
\FOR{$n = 1, \cdots, N$}
\STATE {\bf Step 1} - Let $\boldsymbol{X}_n = \begin{bmatrix} \boldsymbol{x}_{nB} & \cdots & \boldsymbol{x}_{(n+1)B - 1} \end{bmatrix} \in \mathbb{R}^{d \times B}$ collect a sequence of the $i$-th segment.
\STATE {\bf Step 2} - Calculate key, query, values: $\boldsymbol{Q}_n = \boldsymbol{W}_Q \boldsymbol{X}_n$, $\boldsymbol{K}_n = \boldsymbol{W}_K \boldsymbol{X}_n$, and $\boldsymbol{V}_n = \boldsymbol{W}_V \boldsymbol{X}_n $.
\STATE {\bf Step 3} - {Perform attention as: \\
\hspace{3em} $\boldsymbol{O}_n \leftarrow \operatorname{Attn}([\widetilde{\boldsymbol{K}}, \boldsymbol{K}_n], \boldsymbol{Q}_i, [\widetilde{\boldsymbol{V}}, \boldsymbol{V}_n])$}
\IF{$\#(\widetilde{\boldsymbol{K}}, \widetilde{\boldsymbol{V}}) + B \le M$}
\STATE Fill KV cache: $\widetilde{\boldsymbol{K}} \leftarrow [\widetilde{\boldsymbol{K}}, \boldsymbol{K}_n]$,  $\widetilde{\boldsymbol{V}} \leftarrow [\widetilde{\boldsymbol{V}}, \boldsymbol{V}_n]$.
\ELSE
\STATE \algemph{{\bf Step 4} - Compute fusion weights as: \\ \hspace{3em} $\{w_{i,j}\}$ and $\{\widetilde{w}_{i,j}\} \leftarrow$ Equations \ref{eqn:token_weight} and \ref{eqn:token_weight_norm}.}
\STATE \algemph{{\bf Step 5} - Update cached KV pairs: $\forall i \in [M]$,\\
\hspace{3em} $\widetilde{\boldsymbol{k}}_i \leftarrow$  Equation \ref{eqn:token_merge_k}, $\widetilde{\boldsymbol{q}}_i \leftarrow$ Equations \ref{eqn:token_merge_q}.}
\ENDIF
\ENDFOR
\STATE {\bf Return} $\begin{bmatrix}
\boldsymbol{O}_1 & \cdots & \boldsymbol{O}_N
\end{bmatrix}$
\end{algorithmic}
\end{algorithm}

\subsubsection{Complexity Analysis}

With negligible computational overhead, \NAME\, achieves \textbf{constant memory} regardless of sequence length. In Step 3, the attention is computed between a group of $M$ tokens and a group of $B$ tokens.
The memory cost for this step is maintained as $\mathcal{O}(MB)$.
Step 4 synthesizes fusion weights via convolution, whose computation complexity can be as cheap as $\mathcal{O}(L\log L)$ by Fourier transformation.
Afterward, Step 5 leads to a constant-size cache, which guarantees the computation in the next round does not require more memory.
Therefore, the total peak memory cost is $\mathcal{O}(MB + M)$ with an extra $\mathcal{O}(L \log L)$ computation overhead.

\subsubsection{Connection with Token Dropping}

\citet{zhang2023h} proposes to use accumulated attention scores to determine the importance of tokens. The method then auto-regressively keeps tokens with the top scores and discards others. That can be viewed as a special instance of operator $\mathcal{C}$ in Equation~\ref{equ:compression_operator}. However, the heuristic-based method is less expressive compared to our learnable framework.
Specifically, as detailed in Section~\ref{experiment:compression_no_enlength} and Section~\ref{experiment:compression_enlength}, \NAME{}, empowered by the general learnable token compression paradigm, demonstrates superior performance compared to prior arts in token eviction. In addition, our method can be executed on top of other token eviction methods, as to be discussed in Section~\ref{ablation:combination_methods}.

\subsection{Dropping-In Integration of \NAME{}} 
In this section, we introduce how our technique can be easily integrated to pre-trained LLMs, for both long-context inference and long-context training purposes.

\vspace{-0.5em}
\paragraph{Long-Context Efficient Inference} 
Standard LLMs cache all previous KV pairs, resulting in high memory usage that limits their applicability in memory-constrained inference. To address this, we ``drop in" a compressor on top of the pre-trained weights. The compressor is optimized using Algorithm~\ref{alg:lc3_attn} with a minimal fraction of the training data (e.g., 104 million tokens, or 0.0052\% of the 2 trillion tokens used for Llama-2 pre-training~\cite{touvron2023llama}).

During the pre-filling stage, prompts are split into segments of size $B$ before being fed into the LLM. These segments sequentially pass through the LLM, generating and compressing KVs via Equation~\ref{equ:compression_operator}, resulting in compressed KVs of length $M$ that encapsulate the context information. In the generation stage, the segment length is set to 1. Detailed results are provided in Section Section~\ref{experiment:compression_no_enlength}.

As our ``dropping-in" term implies, \textit{the pre-trained weights remain unchanged}, allowing users to switch back to the uncompressed mode simply by removing the compressor heads, when sufficient resources are available for a linearly scaled KV cache.

\vspace{-0.5em}
\paragraph{Long-Context Extension} 
Our method also supports long context extension through post-training tuning, allowing pre-trained LLMs to handle longer contexts without incurring the excessive memory costs. We achieve this by leveraging positional interpolation~\cite{chen2023extending}, inserting compressor heads, and adding LoRA adapters to fine-tune the pre-trained model, following ~\citet{chen2023longlora}'s practice. The fine-tuning procedure is detailed in Algorithm~\ref{alg:lc3_attn}.

\section{Experiment}
We first describe our experimental settings in Sec~\ref{experiment:setting}. Then, we demonstrate our proposed convolutional head as a plug-in tool for pre-trained LLMs, that enables memory-efficient inference, in Sec~\ref{experiment:compression_no_enlength}. Additionally, in Sec~\ref{experiment:compression_enlength}, we apply the proposed method to long context fine-tuning, enabling training with long sequences under fixed-size memory. 

\subsection{Experimental Settings}\label{experiment:setting}
\paragraph{Base Models} 
We select Llama2-7B and Llama2-13B \cite{touvron2023llama} as our base models, each with a maximum context length of 4096 tokens. For inference with context lengths shorter than 4096 tokens (in Sec~\ref{experiment:compression_no_enlength}), we retain the original model weights and fine-tune only the convolutional heads. To extend the context window to 32768 tokens during long context fine-tuning (in Sec~\ref{experiment:compression_enlength}), we utilize positional interpolation for initialization \cite{chen2023extending}.

\paragraph{Convolutional Heads} We insert convolutional heads layer-wise to capture the diverse token relationships across layers. Each convolutional head possesses one layer of 1-D convolutional kernels: its input feature dimension is the dimension of key and value matrices, while the output feature dimension is the target memory size. We set the kernel size to be $21$ by default, as more choices will be validated in Sec~\ref{ablation:different_kernel_size}. Within the same layer, all attention heads would share the same set of convolutional kernel parameters. Thus, for Llama2-7b, a 32-layer model with 256-dimension KV states, we only add  22 million parameters for compressing raw KV states to a memory of 128 tokens.

\begin{figure*}[htb]
    \centering
    \includegraphics[width=1.0\linewidth]{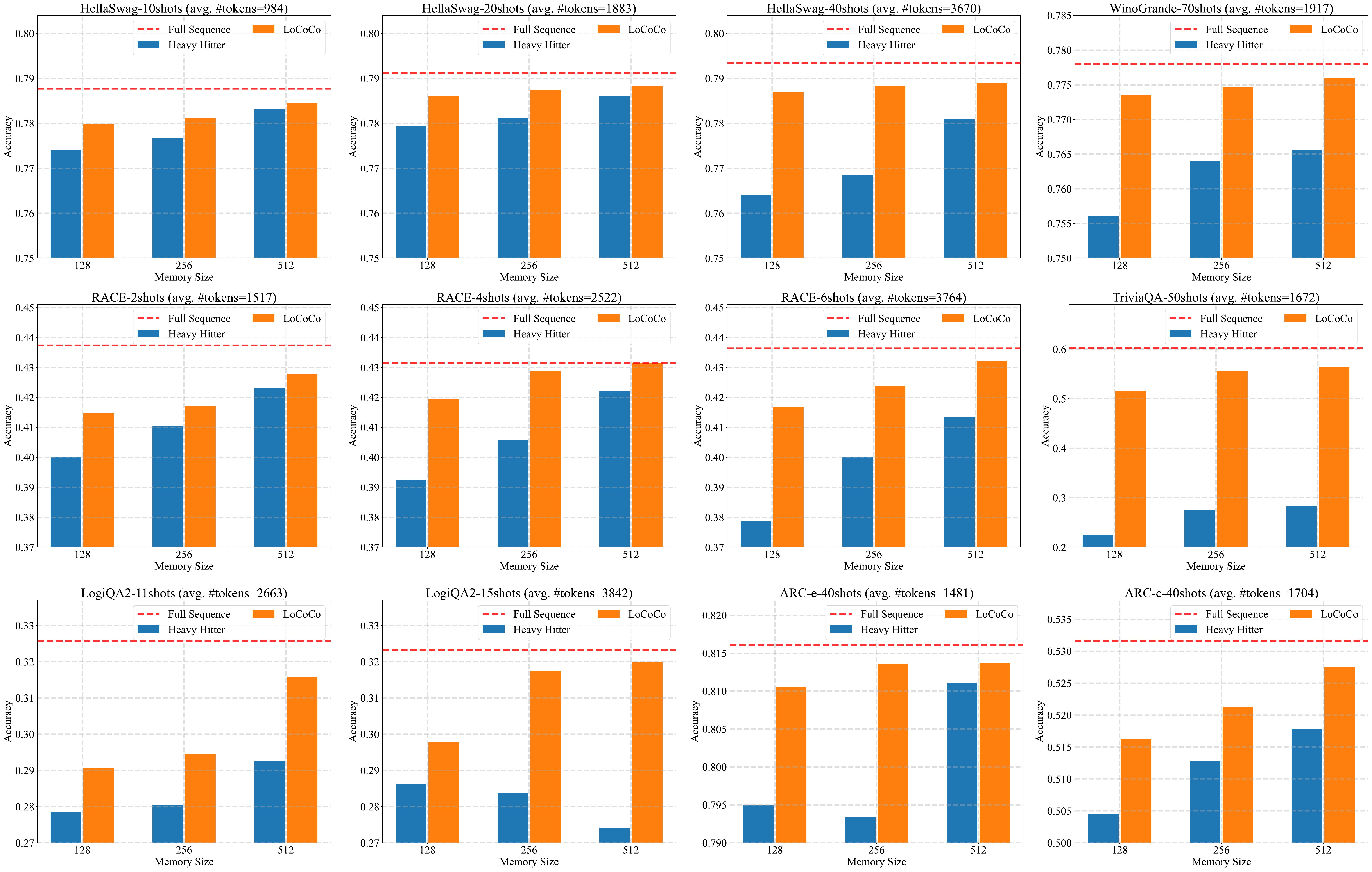}
    \caption{Token merging via convolutional kernels as the drop-in" integration without modifying the original weights. Based on Llama-2-7B \cite{touvron2023llama}, we inserted the convolutional heads on the top of self-attention, and tested the model performance on various few-shot downstream tasks. The input sequence typically consists of about 2000 tokens. We compare our method with \citet{zhang2023h}, a token eviction strategy. We also provide the uncompressed case, where the model uses the full sequence.}
    \label{fig:main_figure1}
\end{figure*}

\begin{table*}[htb]
\centering
\caption{LoCoCo applied to the ChatGLM3-6B-32k \cite{du2021glm} base model, and validated on SCROLLS \cite{shaham2022scrolls}.\label{table:chatglm}}
\resizebox{1.5\columnwidth}{!}{
\begin{tabular}{@{}lcccccc@{}}
\toprule
SCORLLS Task & \multicolumn{1}{l}{QuALITY} & Qasper & SummScreen & GovReport & QMSum & NarrativeQA \\ \midrule
$H_2O$ & 0.4351 & 0.3919 & 0.2498 & 0.3411 & 0.2137 & 0.2433 \\
\rowcolor{Gray}
ours & 0.4689 & 0.4284 & 0.2611 & 0.3617 & 0.2310 & 0.2576 \\
full sequence & 0.4769 & 0.4314 & 0.2636 & 0.3669 & 0.2378 & 0.2605 \\ \bottomrule
\end{tabular}
}
\vspace{-1.5em}
\end{table*}

\begin{table*}[htb]
\centering
\newcolumntype{g}{>{\columncolor{Gray}}l}
\caption{Perplexity evaluated on Proof-Pile-2\cite{azerbayev2023llemma}. We fine-tuned Llama-2-7B \cite{touvron2023llama} to extend the context length from 4K to 8K, 16K, and 32K, respectively. Additionally, we fine-tuned Llama-2-13B, extending the 4K context length to 8K. $T$ denotes the sequence length of the training data, whereas $L$ indicates the chunk size.
\label{table:extend_main}}
\resizebox{1.7\columnwidth}{!}{
\begin{tabular}{@{}llggggggg@{}}
\toprule 
\rowcolor{white}
 & \multicolumn{1}{l}{} &  &  & \multicolumn{5}{c}{Evaluation Context Length} \\ \cmidrule(l){5-9} 
\rowcolor{white}
\multirow{-2}{*}{Size} & \multirow{-2}{*}{\begin{tabular}[c]{@{}l@{}}Training\\ Length $(T)$\end{tabular}} & \multirow{-2}{*}{Method} & \multirow{-2}{*}{\begin{tabular}[r]{@{}l@{}}Attention\\ Complexity\end{tabular}} & \multicolumn{1}{c}{2048} & \multicolumn{1}{c}{4096} & \multicolumn{1}{c}{8192} & \multicolumn{1}{c}{16384} & \multicolumn{1}{r}{32768} \\ \midrule
\rowcolor{white}
 &  & StreamingLLM & $O(L\times (L+8))$ & 4.0373 & 4.0174 & 4.0551 & - & - \\
 \rowcolor{white}
 &  & LongLoRA & $O(L^2)$ & 4.0526 & 3.8111 & 3.6877 & - & - \\
 \rowcolor{white}
 &  & $H_2O$ & $O(L\times (L+512))$  & 3.9653 & 3.7043 & 3.5706 & - & - \\
 &  & Ours & $O(L\times (L+512))$ & 3.9411  & 3.6775  & 3.5414 & - & - \\ 
 \rowcolor{white}
 & \multirow{-4}{*}{8192} & Full Sequence & $O(L\times T)$ & 3.9325 &  3.6558 & 3.5070 & & \\ \cmidrule(lr){2-9}
 \rowcolor{white}
 &  & StreamingLLM & $O(L\times (L+8))$ & 4.0373 & 4.0174 & 4.0551 & 4.0334 & - \\
 \rowcolor{white}
 &  & LongLoRA & $O(L^2)$ & 4.0704 & 3.8125 & 3.6928 & 3.6279 & - \\
 \rowcolor{white}
 &  & $H_2O$ & $O(L\times (L+512))$ & 3.9842 & 3.7173 & 3.5974 & 3.5458 & - \\
 &  & Ours & $O(L\times (L+512))$ & 3.9628 & 3.6958 & 3.5763 & 3.5058 & - \\ 
 \rowcolor{white}
 & \multirow{-4}{*}{16384} & Full Sequence & $O(L\times T)$  & 3.9491 & 3.6619 & 3.5094 & 3.4801 & - \\ \cmidrule(lr){2-9}
 \rowcolor{white}
 &  & StreamingLLM & $O(L\times (L+8))$ & 4.0373 & 4.0174 & 4.0551 & 4.0334 & 4.0171 \\
 \rowcolor{white}
 &  & LongLoRA & $O(L^2)$ & 4.0891 & 3.8348 & 3.7161 & 3.6276 & 3.5916 \\
 \rowcolor{white}
 &  & $H_2O$ & $O(L\times (L+512))$ & 4.0564 & 3.8179 & 3.6570 & 3.5634 & 3.5102 \\
 &  & Ours & $O(L\times (L+512))$ &  4.0253 & 3.8078 & 3.5807 & 3.5145 & 3.4408 \\
 \rowcolor{white}
\multirow{-12}{*}{7b} & \multirow{-4}{*}{32768} & Full Sequence & $O(L\times T)$ & 3.9803 & 3.7703 & 3.5011 & 3.4836 & 3.4012 \\ \midrule
\rowcolor{white}
 &  & StreamingLLM & $O(L\times (L+8))$ & 3.6979 & 3.7013 & 3.7022 & - & - \\
 \rowcolor{white}
 &  & LongLoRA & $O(L^2)$ & 3.7153 & 3.5902 & 3.4511 & - & - \\
 \rowcolor{white}
 &  & $H_2O$ & $O(L\times (L+512))$ & 3.6823 & 3.5482 & 3.4073 & - & - \\
 &  & Ours & $O(L\times (L+512))$ & 3.6798 & 3.4953 & 3.3697 & - & - \\ 
 \rowcolor{white}
 \multirow{-4}{*}{13b} & \multirow{-4}{*}{8192} & Full Sequence & $O(L\times T)$  & 3.6412 & 3.4506 & 3.3421 & - & - \\\bottomrule
\end{tabular}}
\end{table*}

\begin{table*}[]
\centering
\caption{Performance on representative long-context task SCROLLS. \cite{shaham2022scrolls}}
\resizebox{1.5\columnwidth}{!}{
\begin{tabular}{@{}lcccccc@{}}
\toprule
SCORLLS Task & \multicolumn{1}{l}{QuALITY} & Qasper & SummScreen & GovReport & QMSum & NarrativeQA \\ \midrule
LongLoRA & 0.3395 & 0.2421 & 0.1712 & 0.2891 & 0.1792 & 0.1754 \\
$H_2O$ & 0.3461 & 0.2659 & 0.1885 & 0.2924 & 0.1913 & 0.1849 \\
\rowcolor{Gray}
LoCoCo & 0.3528 & 0.2813 & 0.1903 & 0.3113 & 0.2089 & 0.1902 \\
full sequence & 0.3600 & 0.2828 & 0.1945 & 0.3125 & 0.2125 & 0.1942 \\ \bottomrule
\end{tabular}
}
\end{table*}

\paragraph{Compression Details} For post-hoc compression without modifying pre-trained LLMs, we experiment compressing the sequence of length up to $4096$, to fit in the memory size of $128$, $256$, $512$, leading to the compression ratio of $32:1$. 

For experiments on context length extending, we by default set $512$ as the memory size. We will validate more choices ranging from $128$ to $1024$ in Sec~\ref{ablation:different_memory_size}.

\paragraph{Training Details} We use RedPajama \cite{together2023redpajama} as our training dataset. For post-hoc compression experiments, we only tune compression heads for $200$ steps without modifying the pre-trained LLM. For context length extending, we fine-tune the convolutional heads and LoRA adapters (rank 8),  and also allow modifying the embedding and normalization layers, all following \citet{chen2023longlora}. 

For all experiments, we use the learning rates of $5\times10^{-5}$ for LoRA adapters, embedding and normalization layers and $5\times10^{-2}$ for convolutional heads, with linear learning rate schedule. We use the batch size of $128$, and chunk size of $512$. All experiments are run on A6000 (48GB memory) to intentionally test our efficacy with small-memory GPUs, and we use per-device batch size as 1.

\begin{figure*}[htb]
    \centering
    \includegraphics[width=0.8\linewidth]{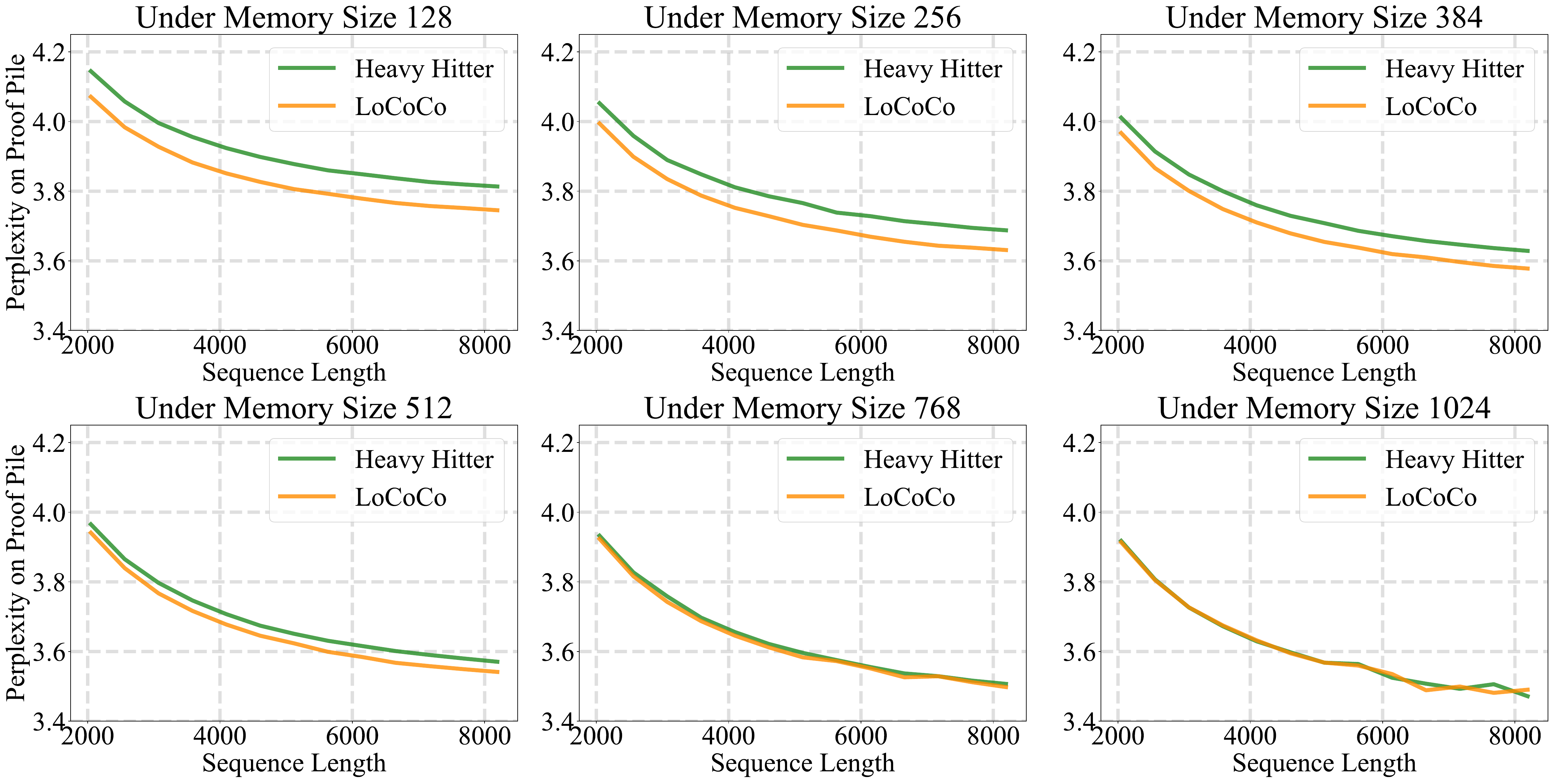}
    \caption{Varying memory sizes during fine-tuning, evaluated on Proof-Pile-2 \cite{azerbayev2023llemma}. Compared to \cite{zhang2023h}, our method shows exceptional performance at large compression ratios, indicating the expressiveness of the merged token.}
    \label{fig:temp2}
\end{figure*}

\subsection{Post-hoc Token Compression of Pre-trained Models}\label{experiment:compression_no_enlength}
At inference, we validate \NAME{} on representative downstream tasks, under target memory sizes varying from $128$ to $512$. We select the reading comprehension dataset RACE \cite{lai2017race} (2, 4, 6 shots), the closed-book question answering dataset TriviaQA \cite{joshi2017triviaqa} (50 shots), and the common sense reasoning dataset: HellaSwag \cite{zellers2019hellaswag} (10, 20, 40 shots), WinoGrande \cite{sakaguchi2021winogrande} (70 shots), and ARC easy and challenge \cite{clark2018think} (40 shots). Note that we deliberately keep the sequence length of each task within the maximum sequence length of the pre-trained Llama-2 \cite{touvron2023llama}.

Using the Llama-2-7b \cite{touvron2023llama} as the base model, we compare our approach with $H_2O$ \cite{zhang2023h}, a recent token dropping method. As in Figure~\ref{fig:main_figure1}, \NAME{} shows exceptional performance on various tasks, especially on tasks whose average sequence length is long. 

\NAME{} be further applied onto any long-context model. We insert convolutional heads on the top of ChatGLM3-6B-32k \cite{du2021glm}, a representative long-context pre-trained model. We evaluate the model on SCROLLS \cite{shaham2022scrolls}, a popular long-context dataset, and Table \ref{table:chatglm} again demonstrates our effectiveness over $H_2O$.

\subsection{Extending Context Length with Limited Memory} \label{experiment:compression_enlength}
In this section, we set the memory size to 512 and extend the pre-trained context length of Llama-2 \cite{touvron2023llama} from 4096 to 8192, 16384, and 32768 using the Red-Pajama pre-training dataset \cite{together2023redpajama}. We conduct experiments on the 7B and 13B models and report perplexity on Proof-Pile-2 \cite{azerbayev2023llemma}. We also validate the model performance under shorter context lengths. 

The results are provided in Table~\ref{table:extend_main}. Besides \citet{zhang2023h}, we also compare with StreamingLLM \cite{xiao2023efficient}, a method handling contexts longer than the pre-trained length in a zero-shot manner. Additionally, we compare with LongLoRA \cite{chen2023longlora}, which utilizes only local tokens without considering global information. Finally, we evaluate the model tuned with uncompressed full sequence length. When combining our proposed token merging with eviction, our method demonstrates superior performance over the aforementioned methods, and shows comparable performance with the uncompressed scenario.

\begin{table}[h]
\label{app:longbench}
\vspace{-1em}
\caption{Evaluation on LongBench~\cite{bai2023longbench}.\label{table:long_bench}}
\centering
\resizebox{0.7\linewidth}{!}{
    \begin{tabular}{l|ccc}\toprule
    Method & LongLoRA & $H_2O$ & LoCoCo \\\midrule
    LongBench & $34.7\%$ & $36.9\%$ & $37.4\%$\\ \bottomrule
    \end{tabular}}
    \vspace{-1.5em}
\end{table}

To further validate our effectiveness, we report our results on LongBench~\cite{bai2023longbench} in Table~\ref{table:long_bench}. We adopt Llama2-13b~\cite{touvron2023llama} and extend the maximum context length to 32K. Compared to LongLoRA~\cite{chen2023longlora} and $H_2O$~\cite{zhang2023h}, our method again achieves superior performance.

\begin{table}
\vspace{-0.5em}
\caption{Comparison on memory usage (during training) and throughput (during inference).\label{table:memory}}
\centering
\resizebox{\linewidth}{!}{
    \begin{tabular}{l|cccc}\toprule
    Method & LongLoRA & $H_2O$& LoCoCo & Full Sequence \\\midrule
    Memory Usage & 49GB & 50GB & 50GB & OOM \\ 
    Throughput (Token/s) & 25 & 32 & 33 & 11 \\\bottomrule
    \end{tabular}}
    \vspace{-1.5em}
\end{table}

\subsection{Memory and Throughput Measurement}
We first test our GPU memory usage during training (tuning): the memory is measured when extending the context length of Llama2-7B to $16$k.
As shown in Table~\ref{table:memory}, performing training directly on the full sequence will exhaust all GPU memory (resulting in ``OOM''). In contrast, our method only requires an additional 1GB of memory compared to LongLoRA~\cite{chen2023longlora} and uses the same amount of memory as $H_2O$~\cite{zhang2023h}. 

We then measure the throughput during inference, at the pre-filling stage. The pre-filling length is set to be $16$k.
As shown in Table~\ref{table:memory}, our method achieves superior throughput compared to all baselines at inference.
For all aforementioned experiments, we set the batch size to 1, and the block size and the KV cache memory size to both $512$. We use Flash Attention v2~\cite{dao2023flashattention} and DeepSpeed Stage 2 by default. The measurements are conducted on the NVIDIA A100 80GB GPU, confirming our inference efficiency.

\section{Ablation}
\subsection{Effectiveness under Different Memory Sizes} \label{ablation:different_memory_size}

We vary the memory size during fine-tuning, ranging from $128$ to $1024$, and compare our method with \cite{zhang2023h}. Based on the pre-trained model Llama-2 \cite{touvron2023llama} whose maximal context length is 4096, we extend it to the length of 8192, on dataset RedPajama \cite{together2023redpajama}. We evaluate the models on Proof-Pile-2 \cite{azerbayev2023llemma} in terms of perplexity. 
As in Figure~\ref{fig:temp2}, our method shows exceptional performance especially at large compression ratios, indicating that \NAME{} could generate more expressive compressed tokens compared to heavy-hitters.

\subsection{Combination with Different Eviction Policies} \label{ablation:combination_methods} 

Our method could either work alone or be integrated with any token eviction policy. 
In Table~\ref{table:abalation:combine}, to extend the maximum context length of Llama-2-7B to 8192 tokens, we showed our core idea of token merging via convolutional heads ($1$) works well alone; ($2$) could be combined with StreamingLLM \cite{xiao2023efficient}, by additionally storing the initial tokens, as known as ``attention sink"; and ($3$) could be further augmented by heavy hitters\cite{zhang2023h}, the ``important tokens" identified by accumulated attention scores. All variants of our methods show superior performance to solely using the previous token eviction method \cite{zhang2023h}. 

\begin{table}[t]
\centering
\vspace{-1em}
\caption{Our method could be performed solely or combined with multiple token eviction methods.\label{table:abalation:combine}}
\resizebox{0.65\columnwidth}{!}{
\begin{tabular}{@{}lc@{}}
\toprule
Method & Perplexity \\ \midrule
$H_2O$ & 3.5714 \\
LoCoCo & 3.5451 \\
LoCoCo w. StreamingLLM & 3.5439 \\
LoCoCo w. $H_2O$ & 3.5414 \\ \bottomrule
\end{tabular}}
\vspace{-1em}
\end{table}

\subsection{Effectiveness under Different Kernel Sizes}\label{ablation:different_kernel_size}
Longer convolutional kernels may also present challenges in optimization. With the Llama-2-7B model \cite{touvron2023llama}, we extend the context length to 8192, employing kernel sizes ranging from $3$ to $21$. We evaluate the fine-tuned model on Proof-Pile-2 \cite{azerbayev2023llemma}, using a context length of 8192. The results are summarized in Table~\ref{table:kernel_size}. We observe stable performance for most size choices, although there are degradations with extremely small kernel sizes. That suggests \NAME{} can work well with moderately sized convolutions, without visible optimization hurdles. 

\begin{table}[htb]
\centering
\vspace{-1em}
\caption{Ablation with different kernel sizes. \label{table:kernel_size}}
\resizebox{\columnwidth}{!}{
\begin{tabular}{@{}ccccccccc@{}}
\toprule
Kernel & \multicolumn{1}{r}{3} & \multicolumn{1}{r}{7} & \multicolumn{1}{r}{17} & \multicolumn{1}{r}{21} & \multicolumn{1}{r}{31} & \multicolumn{1}{r}{41} & \multicolumn{1}{r}{51} & \multicolumn{1}{r}{61} \\ \midrule
PPL & 3.68 & 3.57 & 3.53 & 3.53 & 3.54 & 3.53 & 3.57 & 3.58 \\ \bottomrule
\end{tabular}
}
\vspace{-1em}
\end{table}

\section{Conclusions}
This paper introduces \NAME{}, designed to improve both computation and memory efficiency when dealing with long-context inputs, through the use of a fixed-size KV Cache. We propose a data-driven adaptive token fusion technique, characterized by learnable convolutional kernels. \NAME{} is compatible with any pre-trained Language Models (LLMs), enabling seamless integration with low overhead. Experiments demonstrate that \NAME{} achieves a compression ratio of up to $32:1$ and outperforms baseline methods by up to $27.91\%$ in accuracy. 

\paragraph{Acknowledgements}
Portions of this research were conducted with the advanced computing resources provided by Texas A\&M High Performance Research Computing\footnote{https://hprc.tamu.edu/aces/}. The work is in part supported by the gift funding from \url{https://moffett.ai} (B. Chen) and the National AI Institute for Foundations of Machine Learning  (Z. Wang).

\section*{Impact Statement}
This paper presents work whose goal is to advance the field of efficient and green AI. There are many potential societal consequences of our work, none of which we feel must be specifically highlighted here.

\bibliography{example_paper}

\begin{thebibliography}{66}
\providecommand{\natexlab}[1]{#1}
\providecommand{\url}[1]{\texttt{#1}}
\expandafter\ifx\csname urlstyle\endcsname\relax
  \providecommand{\doi}[1]{doi: #1}\else
  \providecommand{\doi}{doi: \begingroup \urlstyle{rm}\Url}\fi

\bibitem[ntk(2023)]{ntk}
Ntk-aware scaled rope.
\newblock \url{https://www.reddit.com/r/LocalLLaMA/comments/14lz7j5/ntkaware_scaled_rope_allows_llama_models_to_have/}, 2023.

\bibitem[Arora et~al.(2023)Arora, Eyuboglu, Timalsina, Johnson, Poli, Zou, Rudra, and R{\'e}]{arora2023zoology}
Arora, S., Eyuboglu, S., Timalsina, A., Johnson, I., Poli, M., Zou, J., Rudra, A., and R{\'e}, C.
\newblock Zoology: Measuring and improving recall in efficient language models.
\newblock \emph{arXiv preprint arXiv:2312.04927}, 2023.

\bibitem[Azerbayev et~al.(2023)Azerbayev, Schoelkopf, Paster, Santos, McAleer, Jiang, Deng, Biderman, and Welleck]{azerbayev2023llemma}
Azerbayev, Z., Schoelkopf, H., Paster, K., Santos, M.~D., McAleer, S., Jiang, A.~Q., Deng, J., Biderman, S., and Welleck, S.
\newblock Llemma: An open language model for mathematics, 2023.

\bibitem[Bai et~al.(2023)Bai, Lv, Zhang, Lyu, Tang, Huang, Du, Liu, Zeng, Hou, et~al.]{bai2023longbench}
Bai, Y., Lv, X., Zhang, J., Lyu, H., Tang, J., Huang, Z., Du, Z., Liu, X., Zeng, A., Hou, L., et~al.
\newblock Longbench: A bilingual, multitask benchmark for long context understanding.
\newblock \emph{arXiv preprint arXiv:2308.14508}, 2023.

\bibitem[Beltagy et~al.(2020)Beltagy, Peters, and Cohan]{beltagy2020longformer}
Beltagy, I., Peters, M.~E., and Cohan, A.
\newblock Longformer: The long-document transformer.
\newblock \emph{arXiv preprint arXiv:2004.05150}, 2020.

\bibitem[Brown et~al.(2020)Brown, Mann, Ryder, Subbiah, Kaplan, Dhariwal, Neelakantan, Shyam, Sastry, Askell, et~al.]{brown2020language}
Brown, T., Mann, B., Ryder, N., Subbiah, M., Kaplan, J.~D., Dhariwal, P., Neelakantan, A., Shyam, P., Sastry, G., Askell, A., et~al.
\newblock Language models are few-shot learners.
\newblock \emph{Advances in neural information processing systems}, 33:\penalty0 1877--1901, 2020.

\bibitem[Chen et~al.(2021)Chen, Tworek, Jun, Yuan, Pinto, Kaplan, Edwards, Burda, Joseph, Brockman, et~al.]{chen2021evaluating}
Chen, M., Tworek, J., Jun, H., Yuan, Q., Pinto, H. P. d.~O., Kaplan, J., Edwards, H., Burda, Y., Joseph, N., Brockman, G., et~al.
\newblock Evaluating large language models trained on code.
\newblock \emph{arXiv preprint arXiv:2107.03374}, 2021.

\bibitem[Chen et~al.(2023{\natexlab{a}})Chen, Wong, Chen, and Tian]{chen2023extending}
Chen, S., Wong, S., Chen, L., and Tian, Y.
\newblock Extending context window of large language models via positional interpolation.
\newblock \emph{arXiv preprint arXiv:2306.15595}, 2023{\natexlab{a}}.

\bibitem[Chen et~al.(2023{\natexlab{b}})Chen, Qian, Tang, Lai, Liu, Han, and Jia]{chen2023longlora}
Chen, Y., Qian, S., Tang, H., Lai, X., Liu, Z., Han, S., and Jia, J.
\newblock Longlora: Efficient fine-tuning of long-context large language models.
\newblock \emph{arXiv preprint arXiv:2309.12307}, 2023{\natexlab{b}}.

\bibitem[Child et~al.(2019)Child, Gray, Radford, and Sutskever]{child2019generating}
Child, R., Gray, S., Radford, A., and Sutskever, I.
\newblock Generating long sequences with sparse transformers.
\newblock \emph{arXiv preprint arXiv:1904.10509}, 2019.

\bibitem[Choromanski et~al.(2020)Choromanski, Likhosherstov, Dohan, Song, Gane, Sarlos, Hawkins, Davis, Mohiuddin, Kaiser, et~al.]{choromanski2020rethinking}
Choromanski, K., Likhosherstov, V., Dohan, D., Song, X., Gane, A., Sarlos, T., Hawkins, P., Davis, J., Mohiuddin, A., Kaiser, L., et~al.
\newblock Rethinking attention with performers.
\newblock \emph{arXiv preprint arXiv:2009.14794}, 2020.

\bibitem[Clark et~al.(2018)Clark, Cowhey, Etzioni, Khot, Sabharwal, Schoenick, and Tafjord]{clark2018think}
Clark, P., Cowhey, I., Etzioni, O., Khot, T., Sabharwal, A., Schoenick, C., and Tafjord, O.
\newblock Think you have solved question answering? try arc, the ai2 reasoning challenge.
\newblock \emph{arXiv preprint arXiv:1803.05457}, 2018.

\bibitem[Computer(2023)]{together2023redpajama}
Computer, T.
\newblock Redpajama: an open dataset for training large language models, 2023.
\newblock URL \url{https://github.com/togethercomputer/RedPajama-Data}.

\bibitem[Dai et~al.(2019)Dai, Yang, Yang, Carbonell, Le, and Salakhutdinov]{dai2019transformer}
Dai, Z., Yang, Z., Yang, Y., Carbonell, J., Le, Q.~V., and Salakhutdinov, R.
\newblock Transformer-xl: Attentive language models beyond a fixed-length context.
\newblock \emph{arXiv preprint arXiv:1901.02860}, 2019.

\bibitem[Dao(2023)]{dao2023flashattention}
Dao, T.
\newblock Flashattention-2: Faster attention with better parallelism and work partitioning.
\newblock \emph{arXiv preprint arXiv:2307.08691}, 2023.

\bibitem[Dauphin et~al.(2017)Dauphin, Fan, Auli, and Grangier]{dauphin2017language}
Dauphin, Y.~N., Fan, A., Auli, M., and Grangier, D.
\newblock Language modeling with gated convolutional networks.
\newblock In \emph{International conference on machine learning}, pp.\  933--941. PMLR, 2017.

\bibitem[Du et~al.(2021)Du, Qian, Liu, Ding, Qiu, Yang, and Tang]{du2021glm}
Du, Z., Qian, Y., Liu, X., Ding, M., Qiu, J., Yang, Z., and Tang, J.
\newblock Glm: General language model pretraining with autoregressive blank infilling.
\newblock \emph{arXiv preprint arXiv:2103.10360}, 2021.

\bibitem[Fu et~al.(2022)Fu, Dao, Saab, Thomas, Rudra, and R{\'e}]{fu2022hungry}
Fu, D.~Y., Dao, T., Saab, K.~K., Thomas, A.~W., Rudra, A., and R{\'e}, C.
\newblock Hungry hungry hippos: Towards language modeling with state space models.
\newblock \emph{arXiv preprint arXiv:2212.14052}, 2022.

\bibitem[Goyal \& Durrett(2020)Goyal and Durrett]{goyal2020evaluating}
Goyal, T. and Durrett, G.
\newblock Evaluating factuality in generation with dependency-level entailment.
\newblock \emph{arXiv preprint arXiv:2010.05478}, 2020.

\bibitem[Gu \& Dao(2023)Gu and Dao]{gu2023mamba}
Gu, A. and Dao, T.
\newblock Mamba: Linear-time sequence modeling with selective state spaces.
\newblock \emph{arXiv preprint arXiv:2312.00752}, 2023.

\bibitem[Gu et~al.(2020)Gu, Dao, Ermon, Rudra, and R{\'e}]{gu2020hippo}
Gu, A., Dao, T., Ermon, S., Rudra, A., and R{\'e}, C.
\newblock Hippo: Recurrent memory with optimal polynomial projections.
\newblock \emph{Advances in neural information processing systems}, 33:\penalty0 1474--1487, 2020.

\bibitem[Gu et~al.(2021{\natexlab{a}})Gu, Goel, and R{\'e}]{gu2021efficiently}
Gu, A., Goel, K., and R{\'e}, C.
\newblock Efficiently modeling long sequences with structured state spaces.
\newblock \emph{arXiv preprint arXiv:2111.00396}, 2021{\natexlab{a}}.

\bibitem[Gu et~al.(2021{\natexlab{b}})Gu, Johnson, Goel, Saab, Dao, Rudra, and R{\'e}]{gu2021combining}
Gu, A., Johnson, I., Goel, K., Saab, K., Dao, T., Rudra, A., and R{\'e}, C.
\newblock Combining recurrent, convolutional, and continuous-time models with linear state space layers.
\newblock \emph{Advances in neural information processing systems}, 34:\penalty0 572--585, 2021{\natexlab{b}}.

\bibitem[Gu et~al.(2022)Gu, Goel, Gupta, and R{\'e}]{gu2022parameterization}
Gu, A., Goel, K., Gupta, A., and R{\'e}, C.
\newblock On the parameterization and initialization of diagonal state space models.
\newblock \emph{Advances in Neural Information Processing Systems}, 35:\penalty0 35971--35983, 2022.

\bibitem[Gupta et~al.(2022)Gupta, Gu, and Berant]{gupta2022diagonal}
Gupta, A., Gu, A., and Berant, J.
\newblock Diagonal state spaces are as effective as structured state spaces.
\newblock \emph{Advances in Neural Information Processing Systems}, 35:\penalty0 22982--22994, 2022.

\bibitem[Han et~al.(2023)Han, Wang, Xiong, Chen, Ji, and Wang]{han2023lm}
Han, C., Wang, Q., Xiong, W., Chen, Y., Ji, H., and Wang, S.
\newblock Lm-infinite: Simple on-the-fly length generalization for large language models.
\newblock \emph{arXiv preprint arXiv:2308.16137}, 2023.

\bibitem[Hu et~al.(2021)Hu, Shen, Wallis, Allen-Zhu, Li, Wang, Wang, and Chen]{hu2021lora}
Hu, E.~J., Shen, Y., Wallis, P., Allen-Zhu, Z., Li, Y., Wang, S., Wang, L., and Chen, W.
\newblock Lora: Low-rank adaptation of large language models.
\newblock \emph{arXiv preprint arXiv:2106.09685}, 2021.

\bibitem[Jiang et~al.(2023)Jiang, Wu, Luo, Li, Lin, Yang, and Qiu]{jiang2023longllmlingua}
Jiang, H., Wu, Q., Luo, X., Li, D., Lin, C.-Y., Yang, Y., and Qiu, L.
\newblock Longllmlingua: Accelerating and enhancing llms in long context scenarios via prompt compression.
\newblock \emph{arXiv preprint arXiv:2310.06839}, 2023.

\bibitem[Jin et~al.(2024)Jin, Han, Yang, Jiang, Liu, Chang, Chen, and Hu]{jin2024llm}
Jin, H., Han, X., Yang, J., Jiang, Z., Liu, Z., Chang, C.-Y., Chen, H., and Hu, X.
\newblock Llm maybe longlm: Self-extend llm context window without tuning.
\newblock \emph{arXiv preprint arXiv:2401.01325}, 2024.

\bibitem[Joshi et~al.(2017)Joshi, Choi, Weld, and Zettlemoyer]{joshi2017triviaqa}
Joshi, M., Choi, E., Weld, D.~S., and Zettlemoyer, L.
\newblock Triviaqa: A large scale distantly supervised challenge dataset for reading comprehension.
\newblock \emph{arXiv preprint arXiv:1705.03551}, 2017.

\bibitem[Kamalloo et~al.(2023)Kamalloo, Dziri, Clarke, and Rafiei]{kamalloo2023evaluating}
Kamalloo, E., Dziri, N., Clarke, C.~L., and Rafiei, D.
\newblock Evaluating open-domain question answering in the era of large language models.
\newblock \emph{arXiv preprint arXiv:2305.06984}, 2023.

\bibitem[Katharopoulos et~al.(2020)Katharopoulos, Vyas, Pappas, and Fleuret]{katharopoulos2020transformers}
Katharopoulos, A., Vyas, A., Pappas, N., and Fleuret, F.
\newblock Transformers are rnns: Fast autoregressive transformers with linear attention.
\newblock In \emph{International conference on machine learning}, pp.\  5156--5165. PMLR, 2020.

\bibitem[Kim(2014)]{kim2014convolutional}
Kim, Y.
\newblock Convolutional neural networks for sentence classification.
\newblock \emph{arXiv preprint arXiv:1408.5882}, 2014.

\bibitem[Kitaev et~al.(2020)Kitaev, Kaiser, and Levskaya]{kitaev2020reformer}
Kitaev, N., Kaiser, {\L}., and Levskaya, A.
\newblock Reformer: The efficient transformer.
\newblock \emph{arXiv preprint arXiv:2001.04451}, 2020.

\bibitem[Lai et~al.(2017)Lai, Xie, Liu, Yang, and Hovy]{lai2017race}
Lai, G., Xie, Q., Liu, H., Yang, Y., and Hovy, E.
\newblock Race: Large-scale reading comprehension dataset from examinations.
\newblock \emph{arXiv preprint arXiv:1704.04683}, 2017.

\bibitem[Lewkowycz et~al.(2022)Lewkowycz, Andreassen, Dohan, Dyer, Michalewski, Ramasesh, Slone, Anil, Schlag, Gutman-Solo, et~al.]{lewkowycz2022solving}
Lewkowycz, A., Andreassen, A., Dohan, D., Dyer, E., Michalewski, H., Ramasesh, V., Slone, A., Anil, C., Schlag, I., Gutman-Solo, T., et~al.
\newblock Solving quantitative reasoning problems with language models.
\newblock \emph{Advances in Neural Information Processing Systems}, 35:\penalty0 3843--3857, 2022.

\bibitem[Li et~al.(2022)Li, Choi, Chung, Kushman, Schrittwieser, Leblond, Eccles, Keeling, Gimeno, Dal~Lago, et~al.]{li2022competition}
Li, Y., Choi, D., Chung, J., Kushman, N., Schrittwieser, J., Leblond, R., Eccles, T., Keeling, J., Gimeno, F., Dal~Lago, A., et~al.
\newblock Competition-level code generation with alphacode.
\newblock \emph{Science}, 378\penalty0 (6624):\penalty0 1092--1097, 2022.

\bibitem[Liu et~al.(2023)Liu, Lin, Hewitt, Paranjape, Bevilacqua, Petroni, and Liang]{liu2023lost}
Liu, N.~F., Lin, K., Hewitt, J., Paranjape, A., Bevilacqua, M., Petroni, F., and Liang, P.
\newblock Lost in the middle: How language models use long contexts.
\newblock \emph{arXiv preprint arXiv:2307.03172}, 2023.

\bibitem[Massaroli et~al.(2023)Massaroli, Poli, Fu, Kumbong, Parnichkun, Timalsina, Romero, McIntyre, Chen, Rudra, et~al.]{massaroli2023laughing}
Massaroli, S., Poli, M., Fu, D.~Y., Kumbong, H., Parnichkun, R.~N., Timalsina, A., Romero, D.~W., McIntyre, Q., Chen, B., Rudra, A., et~al.
\newblock Laughing hyena distillery: Extracting compact recurrences from convolutions.
\newblock \emph{arXiv preprint arXiv:2310.18780}, 2023.

\bibitem[Mohtashami \& Jaggi(2023{\natexlab{a}})Mohtashami and Jaggi]{mohtashami2023landmark}
Mohtashami, A. and Jaggi, M.
\newblock Landmark attention: Random-access infinite context length for transformers, 2023{\natexlab{a}}.

\bibitem[Mohtashami \& Jaggi(2023{\natexlab{b}})Mohtashami and Jaggi]{mohtashami2023random}
Mohtashami, A. and Jaggi, M.
\newblock Random-access infinite context length for transformers.
\newblock In \emph{Thirty-seventh Conference on Neural Information Processing Systems}, 2023{\natexlab{b}}.

\bibitem[Peng et~al.(2023)Peng, Quesnelle, Fan, and Shippole]{peng2023yarn}
Peng, B., Quesnelle, J., Fan, H., and Shippole, E.
\newblock Yarn: Efficient context window extension of large language models.
\newblock \emph{arXiv preprint arXiv:2309.00071}, 2023.

\bibitem[Poli et~al.(2023)Poli, Massaroli, Nguyen, Fu, Dao, Baccus, Bengio, Ermon, and R{\'e}]{poli2023hyena}
Poli, M., Massaroli, S., Nguyen, E., Fu, D.~Y., Dao, T., Baccus, S., Bengio, Y., Ermon, S., and R{\'e}, C.
\newblock Hyena hierarchy: Towards larger convolutional language models.
\newblock \emph{arXiv preprint arXiv:2302.10866}, 2023.

\bibitem[Press et~al.(2021)Press, Smith, and Lewis]{press2021train}
Press, O., Smith, N.~A., and Lewis, M.
\newblock Train short, test long: Attention with linear biases enables input length extrapolation.
\newblock \emph{arXiv preprint arXiv:2108.12409}, 2021.

\bibitem[Radford et~al.(2018)Radford, Narasimhan, Salimans, Sutskever, et~al.]{radford2018improving}
Radford, A., Narasimhan, K., Salimans, T., Sutskever, I., et~al.
\newblock Improving language understanding by generative pre-training.
\newblock 2018.

\bibitem[Radford et~al.(2019)Radford, Wu, Child, Luan, Amodei, Sutskever, et~al.]{radford2019language}
Radford, A., Wu, J., Child, R., Luan, D., Amodei, D., Sutskever, I., et~al.
\newblock Language models are unsupervised multitask learners.
\newblock \emph{OpenAI blog}, 1\penalty0 (8):\penalty0 9, 2019.

\bibitem[Ribar et~al.(2023)Ribar, Chelombiev, Hudlass-Galley, Blake, Luschi, and Orr]{ribar2023sparq}
Ribar, L., Chelombiev, I., Hudlass-Galley, L., Blake, C., Luschi, C., and Orr, D.
\newblock Sparq attention: Bandwidth-efficient llm inference.
\newblock \emph{arXiv preprint arXiv:2312.04985}, 2023.

\bibitem[Roy et~al.(2021)Roy, Saffar, Vaswani, and Grangier]{roy2021efficient}
Roy, A., Saffar, M., Vaswani, A., and Grangier, D.
\newblock Efficient content-based sparse attention with routing transformers.
\newblock \emph{Transactions of the Association for Computational Linguistics}, 9:\penalty0 53--68, 2021.

\bibitem[Sakaguchi et~al.(2021)Sakaguchi, Bras, Bhagavatula, and Choi]{sakaguchi2021winogrande}
Sakaguchi, K., Bras, R.~L., Bhagavatula, C., and Choi, Y.
\newblock Winogrande: An adversarial winograd schema challenge at scale.
\newblock \emph{Communications of the ACM}, 64\penalty0 (9):\penalty0 99--106, 2021.

\bibitem[Shaham et~al.(2022)Shaham, Segal, Ivgi, Efrat, Yoran, Haviv, Gupta, Xiong, Geva, Berant, et~al.]{shaham2022scrolls}
Shaham, U., Segal, E., Ivgi, M., Efrat, A., Yoran, O., Haviv, A., Gupta, A., Xiong, W., Geva, M., Berant, J., et~al.
\newblock Scrolls: Standardized comparison over long language sequences.
\newblock \emph{arXiv preprint arXiv:2201.03533}, 2022.

\bibitem[Tay et~al.(2022)Tay, Dehghani, Bahri, and Metzler]{tay2022efficient}
Tay, Y., Dehghani, M., Bahri, D., and Metzler, D.
\newblock Efficient transformers: A survey.
\newblock \emph{ACM Computing Surveys}, 55\penalty0 (6):\penalty0 1--28, 2022.

\bibitem[Touvron et~al.(2023)Touvron, Martin, Stone, Albert, Almahairi, Babaei, Bashlykov, Batra, Bhargava, Bhosale, et~al.]{touvron2023llama}
Touvron, H., Martin, L., Stone, K., Albert, P., Almahairi, A., Babaei, Y., Bashlykov, N., Batra, S., Bhargava, P., Bhosale, S., et~al.
\newblock Llama 2: Open foundation and fine-tuned chat models.
\newblock \emph{arXiv preprint arXiv:2307.09288}, 2023.

\bibitem[Tworkowski et~al.(2023)Tworkowski, Staniszewski, Pacek, Wu, Michalewski, and Mi{\l}o{\'s}]{tworkowski2023focused}
Tworkowski, S., Staniszewski, K., Pacek, M., Wu, Y., Michalewski, H., and Mi{\l}o{\'s}, P.
\newblock Focused transformer: Contrastive training for context scaling.
\newblock \emph{arXiv preprint arXiv:2307.03170}, 2023.

\bibitem[Vaswani et~al.(2017)Vaswani, Shazeer, Parmar, Uszkoreit, Jones, Gomez, Kaiser, and Polosukhin]{vaswani2017attention}
Vaswani, A., Shazeer, N., Parmar, N., Uszkoreit, J., Jones, L., Gomez, A.~N., Kaiser, {\L}., and Polosukhin, I.
\newblock Attention is all you need.
\newblock \emph{Advances in neural information processing systems}, 30, 2017.

\bibitem[Wang et~al.(2020)Wang, Li, Khabsa, Fang, and Ma]{wang2020linformer}
Wang, S., Li, B.~Z., Khabsa, M., Fang, H., and Ma, H.
\newblock Linformer: Self-attention with linear complexity.
\newblock \emph{arXiv preprint arXiv:2006.04768}, 2020.

\bibitem[Wu et~al.(2022)Wu, Rabe, Hutchins, and Szegedy]{wu2022memorizing}
Wu, Y., Rabe, M.~N., Hutchins, D., and Szegedy, C.
\newblock Memorizing transformers.
\newblock \emph{arXiv preprint arXiv:2203.08913}, 2022.

\bibitem[Xiao et~al.(2023)Xiao, Tian, Chen, Han, and Lewis]{xiao2023efficient}
Xiao, G., Tian, Y., Chen, B., Han, S., and Lewis, M.
\newblock Efficient streaming language models with attention sinks.
\newblock \emph{arXiv preprint arXiv:2309.17453}, 2023.

\bibitem[Xiong et~al.(2023)Xiong, Liu, Molybog, Zhang, Bhargava, Hou, Martin, Rungta, Sankararaman, Oguz, et~al.]{xiong2023effective}
Xiong, W., Liu, J., Molybog, I., Zhang, H., Bhargava, P., Hou, R., Martin, L., Rungta, R., Sankararaman, K.~A., Oguz, B., et~al.
\newblock Effective long-context scaling of foundation models.
\newblock \emph{arXiv preprint arXiv:2309.16039}, 2023.

\bibitem[Xiong et~al.(2021)Xiong, Zeng, Chakraborty, Tan, Fung, Li, and Singh]{xiong2021nystromformer}
Xiong, Y., Zeng, Z., Chakraborty, R., Tan, M., Fung, G., Li, Y., and Singh, V.
\newblock Nystr{\"o}mformer: A nystr{\"o}m-based algorithm for approximating self-attention.
\newblock In \emph{Proceedings of the AAAI Conference on Artificial Intelligence}, volume~35, pp.\  14138--14148, 2021.

\bibitem[Yuan et~al.(2022)Yuan, Coenen, Reif, and Ippolito]{yuan2022wordcraft}
Yuan, A., Coenen, A., Reif, E., and Ippolito, D.
\newblock Wordcraft: story writing with large language models.
\newblock In \emph{27th International Conference on Intelligent User Interfaces}, pp.\  841--852, 2022.

\bibitem[Zaheer et~al.(2020)Zaheer, Guruganesh, Dubey, Ainslie, Alberti, Ontanon, Pham, Ravula, Wang, Yang, et~al.]{zaheer2020big}
Zaheer, M., Guruganesh, G., Dubey, K.~A., Ainslie, J., Alberti, C., Ontanon, S., Pham, P., Ravula, A., Wang, Q., Yang, L., et~al.
\newblock Big bird: Transformers for longer sequences.
\newblock \emph{Advances in neural information processing systems}, 33:\penalty0 17283--17297, 2020.

\bibitem[Zellers et~al.(2019)Zellers, Holtzman, Bisk, Farhadi, and Choi]{zellers2019hellaswag}
Zellers, R., Holtzman, A., Bisk, Y., Farhadi, A., and Choi, Y.
\newblock Hellaswag: Can a machine really finish your sentence?
\newblock \emph{arXiv preprint arXiv:1905.07830}, 2019.

\bibitem[Zhang et~al.(2023{\natexlab{a}})Zhang, Ram, Hawkins, Zha, and Zhao]{zhang2023efficient}
Zhang, Q., Ram, D., Hawkins, C., Zha, S., and Zhao, T.
\newblock Efficient long-range transformers: You need to attend more, but not necessarily at every layer.
\newblock \emph{arXiv preprint arXiv:2310.12442}, 2023{\natexlab{a}}.

\bibitem[Zhang et~al.(2024)Zhang, Li, and Liu]{zhang2024extending}
Zhang, Y., Li, J., and Liu, P.
\newblock Extending llms' context window with 100 samples.
\newblock \emph{arXiv preprint arXiv:2401.07004}, 2024.

\bibitem[Zhang et~al.(2023{\natexlab{b}})Zhang, Sheng, Zhou, Chen, Zheng, Cai, Song, Tian, R{\'e}, Barrett, et~al.]{zhang2023h}
Zhang, Z., Sheng, Y., Zhou, T., Chen, T., Zheng, L., Cai, R., Song, Z., Tian, Y., R{\'e}, C., Barrett, C., et~al.
\newblock H $ \_2 $ o: Heavy-hitter oracle for efficient generative inference of large language models.
\newblock \emph{arXiv preprint arXiv:2306.14048}, 2023{\natexlab{b}}.

\bibitem[Zhu et~al.(2023)Zhu, Yang, Wang, Song, Wu, Wei, and Li]{zhu2023pose}
Zhu, D., Yang, N., Wang, L., Song, Y., Wu, W., Wei, F., and Li, S.
\newblock Pose: Efficient context window extension of llms via positional skip-wise training.
\newblock \emph{arXiv preprint arXiv:2309.10400}, 2023.

\end{thebibliography}
\bibliographystyle{icml2024}



\end{document}